%% file: main.tex
\documentclass[a4paper, 11pt]{article}
\usepackage[utf8]{inputenc} 
\usepackage[T1]{fontenc}
\usepackage{lmodern}
\usepackage{graphicx}
\usepackage{caption}
\usepackage{subcaption}
\usepackage{booktabs}
\usepackage{plain}
\usepackage{pbox}
\usepackage{xcolor,soul}
\usepackage[colorlinks]{hyperref}
\usepackage{multirow}
\usepackage{rotating}
\usepackage{amssymb}
\usepackage{authblk}

\usepackage{geometry}
\title{Physical Activity Recognition Based on a Parallel Approach for an Ensemble of Machine Learning and Deep Learning Classifiers} 
\author[1, 2]{\large M. Abid}
\author[1, 3]{\large A. Khabou}
\author[1, 2]{\large Y. Ouakrim}
\author[1,4]{\large H. Watel}
\author[5,2]{\large S. Chemkhi}
\author[3]{\large A. Mitiche}
\author[5]{\large A. Benazza-Benyahia}
\author[1, 2]{\large N. Mezghani}
\affil[1]{\footnotesize Laboratoire LIO, Centre de Recherche du CHUM, Montreal, Quebec, Canada} 
\affil[2]{\footnotesize LICEF institute, TELUQ university, Montreal, Quebec, Canada} 
\affil[3]{\footnotesize INRS, Centre \'{E}nergie, Mat\'{e}riaux et T\'{e}l\'{e}communications, Montreal, Canada} 
\affil[4]{\footnotesize \'{E}cole de Technologie Sup\'{e}rieure (\'{E}TS), Montreal, Canada}
\affil[5]{\footnotesize University of Carthage SUP'COM, LR11TIC01, COSIM Lab., 2083, El Ghazala, Tunisia}
 \date{}
\begin{document}
\maketitle
\abstract{ 
\noindent
Human activity recognition (HAR) by wearable sensor devices embedded in the Internet of things (IOT) can play a significant role in remote health monitoring and emergency notification, to provide healthcare of higher standards. The purpose of this study is to investigate a  human activity recognition method of accrued decision accuracy and speed of execution to be applicable in healthcare. This method classifies wearable sensor acceleration time series data of human movement using efficient classifier combination of feature engineering-based and feature learning-based data representation. Leave-one-subject-out cross-validation of the method with data acquired from $44$ subjects wearing a single waist-worn accelerometer on a smart textile, and engaged in a variety of $10$ activities, yields  an average recognition rate of $90\%$, performing significantly better than individual classifiers. The method easily accommodates   functional and computational parallelization  to bring execution time significantly down.

\textbf{Keywords:} machine learning; deep learning; big data; data streams; Internet of Things; sensor data; intelligent systems: multivariate time series; tensor.}
\input{Introduction}
\input{Materials}
\input{Methods}
\input{Results}

\input{Conclusion}

\input{Acknowledgment}
\bibliographystyle{plain}
\bibliography{references}

\end{document}

%% file: introduction.tex
\section{Introduction}
Miniaturization of complex electrical devices at continually lower cost, has brought about the development of a variety of wearable sensors and their embedding in healthcare-dedicated Internet of things (IoT). The broad purpose of a healthcare IoT, sometimes called Internet of medical things, abbreviated IoMT, is to provide a network of embedded systems to acquire, communicate, and analyze data for remote medical practice of accrued quality. Sensors in the embedded systems of IoMT can perform a variety of useful measurements, such as heart rate, body temperature, blood pressure, temporal data, as with electrocardiography (ECG), and activity data, such as movement acceleration. 
The purpose of this study is to investigate a  human activity recognition (HAR) method of accrued decision accuracy and speed of execution to be applicable and practicable in healthcare IoT applications. The method uses acceleration data of human movement recorded by a single, comfortably worn accelerometer. Recent HAR research  in healthcare has used various wearable sensors for health monitoring and physical rehabilitation systems~\cite{Acampora2013}. Such systems aim at developing methods for automatically recognizing human physical activities by analyzing data gathered by sensors in wearable devices.
The basic problem is to assign a time-series segment of sensor data to a corresponding activity during that time segment~\cite{Thomas2011}. \\

In general, a HAR system is composed of the following steps: (1) Sensor data time-series acquisition, (2) raw  pre-processing, (3 time-series segmentation, (4) feature engineering or learning, and feature selection, and (5) classification. 
First, wearable sensors collect time series data from users. Then, the raw data collected is processed and represented in the form of labeled multivariate time series. Segmentation is the process of dividing the time series into smaller data segments~\cite{Banos2014}. Most of the segmentation techniques can be divided into three categories: activity-defined windowing, event-defined windowing, and sliding windowing. The sliding window strategy, which allows a degree of overlap between fixed-size windows, has been prevalent. Feature engineering uses domain knowledge, and other prior information, to define the features of data representation~\cite{Bengio2013}, and feature learning determines a mapping from the data domain to the feature domain, and this is generally done by neural networks. Current research has not clearly established which of feature engineering and feature learning is more potent in HAR. The recent survey~\cite{Figo2010} discusses several  feature engineering schemes for accelerometer data in HAR, including spectral analysis by Fast Fourier Transforms (FFT), statistics-based metrics, and string matching. \\
 Machine learning can model a wide range of human physical activities for HAR in wearable-sensor embedded systems. However, serious challenges remain. First, both training and learning technique evaluation require large annotated data sets. This can be both a data-intensive and computation-intensive process. Thus, it is important to design parallel algorithms that fully exploit the computational capacity of the target machine, and reduce the training time. Moreover, technical issues such as parallel ensemble learning algorithms that aim at optimizing both the accuracy and computational costs are not yet fully addressed by previous research works.\\
There have been several HAR studies using wearable sensors, in which various human activities were investigated, including common daily activities, ambulation, and fitness exercise, using various sensors, including smart watches, to collect data for machine learning algorithms to process ~\cite{Bao2004, Lara2013, Chen2020, Attal2015, Esch2013}. Expectedly so, most of the sensor measurements these studies used, such as GPS data, heart rate, and motion acceleration, relate to the wearer movement, but motion acceleration has had the strongest impact~\cite{Staudenmayer2009, Ravi2005, Cheung2011}. These are collected typically from a small number of subjects and in restrictive settings~\cite{Bao2004,Lara2013}). In addition, it is not uncommon that several sensors had to be inconveniently worn. 
The results of current research firmly establish the merit and feasibility of wearable sensor HAR, and justify further investigation to develop practicable, parallel efficient algorithms. The purpose of this study is to investigate a parallel classifier combination method toward this end. Specifically, the contributions of this study are:
\begin{itemize}
   \item A large dataset to serve HAR sytem development, recorded on  participants using a comfortable smart textile garment with an embedded single waist-worn accelerometer. 
   \item A parallel architecture to combine traditional and deep learning pattern classification algorithms, for accrued computational and classification accuracy, that we refereed to as an ensemble learning architecture.
   This architecture includes both the training and testing aspects of algorithm development, for ease of application development.
   \item A parallel implementation of this ensemble learning architecture.
\end{itemize}
The remainder of this paper is organized as follows: Section~\ref{material} presents a detailed description of the materials and methods used in this study, which includes data acquisition, preprocessing, and the proposed parallel architecture framework.
Section~\ref{results} describes and discusses the experimental results. Finally, Section~\ref{conclusion} contains a conclusion and perspectives for further research. 

%% file: Materials.tex
\section{Materials and Methods}
\label{material}
\subsection{Data acquisition}
\paragraph{Participants}
Data acquisition is performed in two stages. At the first stage, a convenience sample of 14 healthy and young volunteers (age $25.43 \pm 7.51$ years, weight $60.7 \pm 6.7$ kg, and height $172.7 \pm 7.2$ cm) is retained to participate in the data acquisition. At the second stage, 30 healthy and young volunteers (age $24.26 \pm 3.35$ years, weight $68.4 \pm 12.3$ kg, and height $170.3 \pm 0.08$ cm) participate in the data acquisition. 
The two merged datasets will be the input of the HAR system for healthcare. 
The study is approved by the Ethic committee (ethical approval code: RDCPJ 522843-17) of the Centre Hospitalier de l'Universit\'{e} de Montr\'{e}al (CHUM), the \'{E}cole de Technologie Sup\'{e}rieure (\'{E}TS), and the TÉLUQ university, Canada. Data collection is performed in the biomechanics laboratory on the 7th floor of the research center of CHUM (CRCHUM). Informed consent was obtained from all subjects involved in the study. \\
\paragraph{Equipment}
Data are acquired with a single waist-mounted three-axial accelerometer (a non-invasive sensor with a $13$-bit resolution and a frequency of $64$ Hz) embedded in a health monitoring wearable shirt, Hexoskin (Carré Technologies Inc. in Montréal, Canada). The latter includes also three other sensors to record cardiac and respiratory data. These sensor data are not used in the present study. The sleeveless shirt is made of a stretchable fabric ($73\%$ micro polyamide and $27\%$ elastane), with anti-bacterial, breathable, lightweight, UV-protective and quick-dry properties. Thus, it is easy to put on, comfortable and can be used in any ambient environment. The accelerometer data acquisition can be performed continuously without hampering the movements of the person wearing it. When in use, the recording device connector slot is plugged into the shirt connector. Once connected, the accelerometer data are transmitted from the recording device to Hexoskin servers via Bluetooth and a dedicated cell phone application (the Hexoskin app), that outputs a report capturing health-related information of the performer. The data can be downloaded later to a PC via a USB cable and the HxServices software. The validity of the Hexoskin as an accelerometer based physical activity monitor has been already approved~\cite{Banerjee2015, Cherif2018}.
\paragraph{Research protocol}
\label{research_protocol}
The dataset used in this study is acquired by our research team. It  will serve as a basis for the development of a human physical activity recognition systems for medical purposes. In the experiments, $44$ participants wearing the Hexoskin shirt, perform $10$ activities. Before the acquisition, the participant anthropometrics are recorded (i.e., age, gender, height, weight). 
In order to capture the intra-subject variability and collect a big dataset, each subject was asked to perform $6$ trials for each sequence of activities. That is, the participant performs the different activities in a different order in each trial.
A sequence of the activities performed in the laboratory environment is: go up the stairs two floors (A1), go down the stairs two floors (A2), walk for $30$ seconds along the corridor (A3), run for $30$ seconds along the same corridor (A4), sit for $30$ seconds on a chair (A5), fall-right, -left, -front and -black (respectively A6 to A9), and lie intentionally on a mattress (A10).

The $11^{th}$ class corresponds to transitional activities. This class is removed from the dataset before training.\\
We note that trials are video recorded using a separate smartphone. Then, the video is used as a benchmark to annotate the activities. Each video is tagged by a timestamp in milliseconds.
\subsection{Preprocessing}
\begin{figure}
\centering
\begin{subfigure}{\textwidth}
\centering
\includegraphics[width=5in]{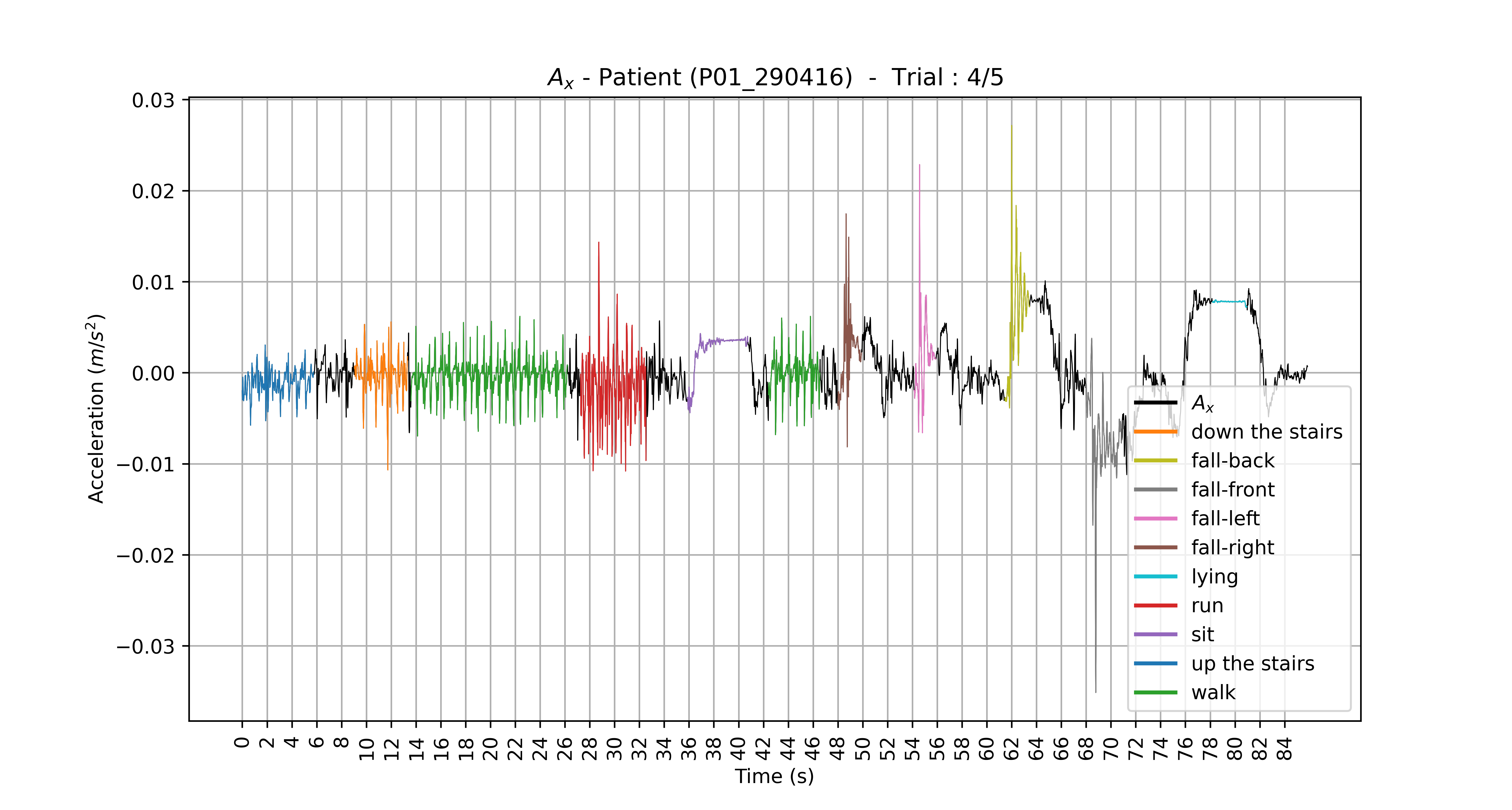}
\caption{Raw x-axis labeled accelerometer data.}
\end{subfigure}
\begin{subfigure}{\textwidth}
\centering
\includegraphics[width=5in]{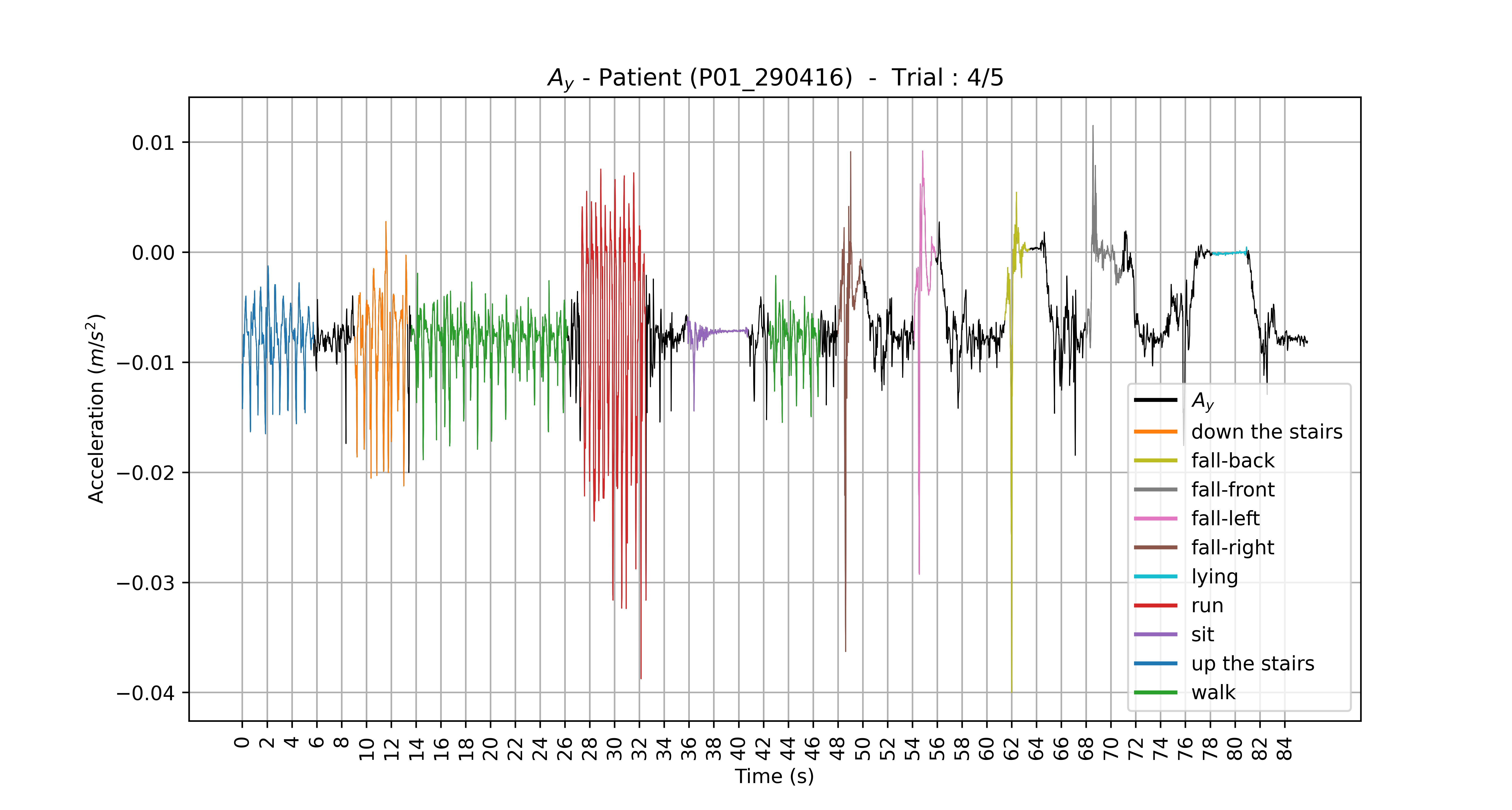}
\caption{Raw y-axis labeled accelerometer data.}
\end{subfigure}
\begin{subfigure}{\textwidth}
\centering
\includegraphics[width=5in]{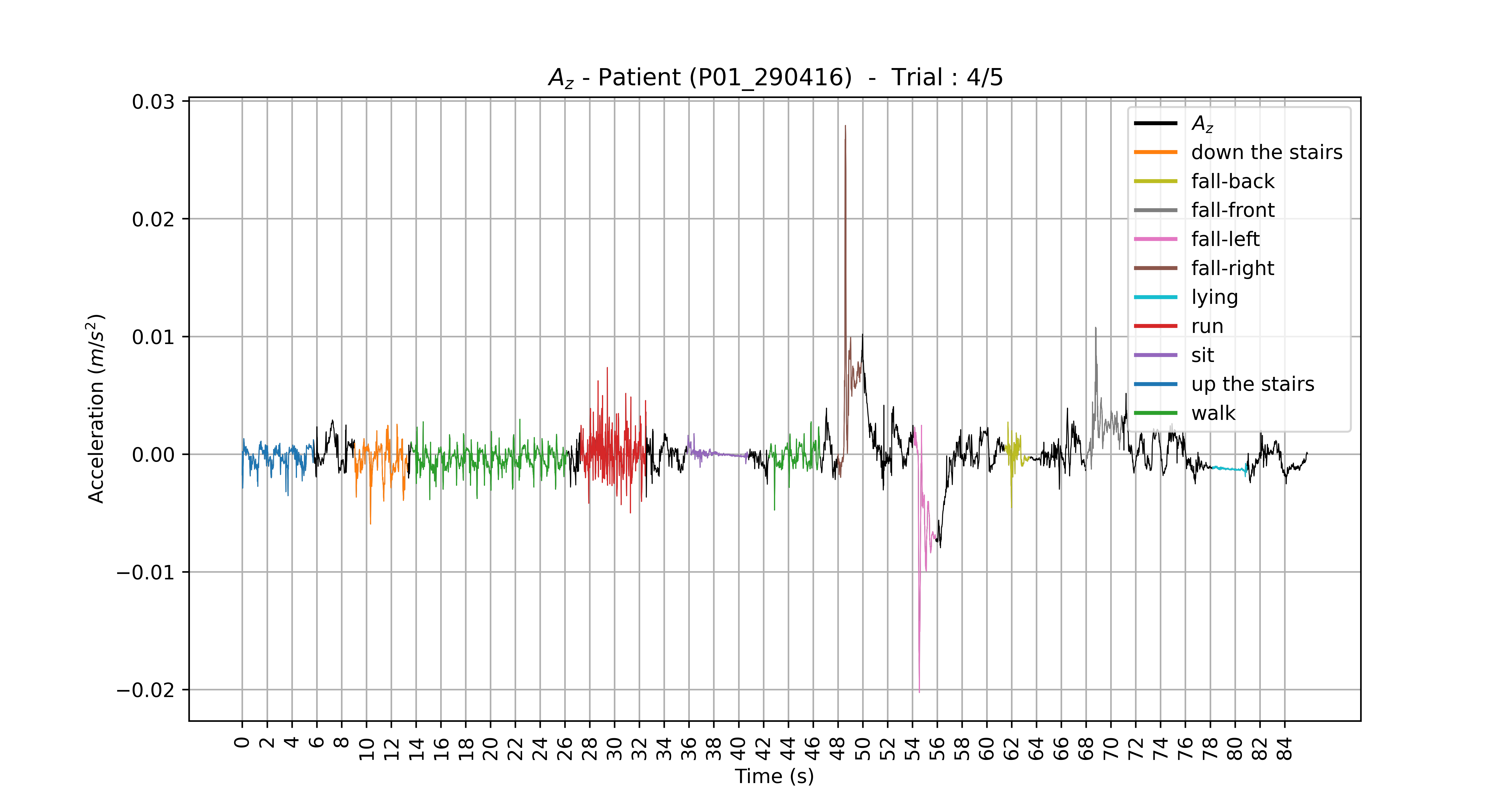}
\caption{Raw z-axis labeled accelerometer data.}
\end{subfigure}
\caption{Plots of the raw three-axis labeled accelerometer data corresponding to one trial of a given participant (P01\_290416). Each data cluster corresponding to an activity label has a different color on the plot.\label{xyzplots}}
\end{figure}

\paragraph{Labeling procedure} First, two members of our research team analyze each video. They manually and precisely annotate the temporal bounds of the observed activities (i.e. start and end times of each activity) in the video. Hence, a ground truth csv file is created for each video with the start and end times of the activities, as well as the class labels. Then, two other members of our research team verify the annotation process. The raw unlabeled accelerometer records of each axis are downloaded separately as a .wav file from the Hexoskin server. The first column represents the timestamp (in seconds) and the second column represents the data itself. We retrieve the start date and time of the data acquisition from the "statistics.csv" file downloaded from the Hexoskin server. 
A second csv file is then automatically generated. In this file, the first column represents the timestamp (in milliseconds). Data from the $x$, $y$, and $z$ axis are respectively in the 2nd, 3rd and 4th column. The Euclidean norm is automatically computed and stored in the 5th column. Then, we automatically assign a ground truth label to each timestamp of the data stream based on the ground truth csv file. To fine-tune the labeling procedure, we visually inspect the plots of the raw three-axis labeled accelerometer data. Each data cluster corresponding to an activity label has a different color on the plot as depicted in Figure~\ref{xyzplots}. 
The raw data are processed using Matlab R2019a. All the raw and processed data are anonymized and stored in the server of the Imaging and Orthopaedics Research Laboratory (LIO), which is affiliated with the \'{E}TS and CRCHUM. The data presented in this study are available on request from the corresponding author. The data are not publicly available due to ethical and legal restrictions on consent given by research participants.\\

\begin{figure}
    \centering
    \includegraphics[width=5in]{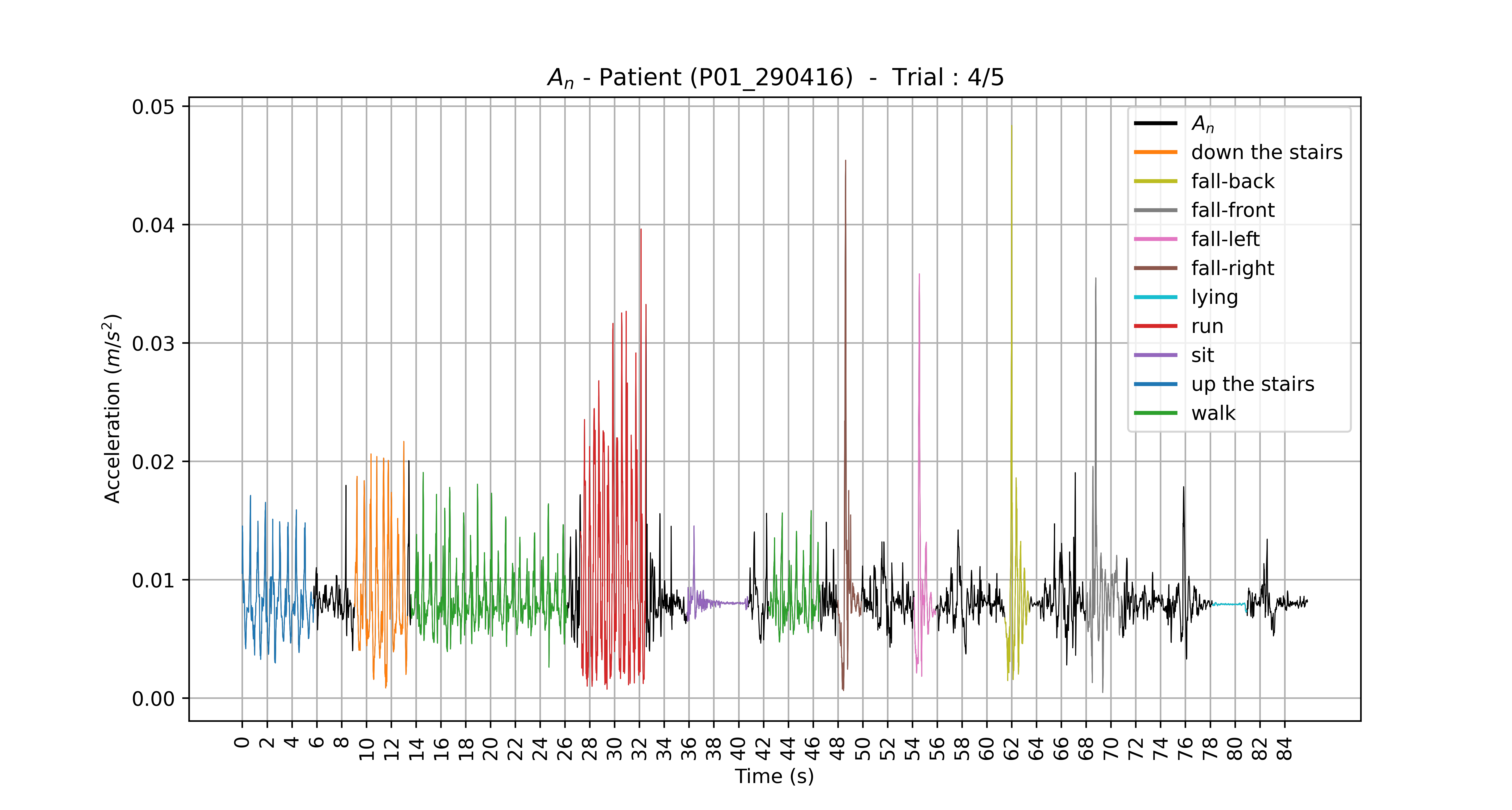}
    \caption{Plot of the raw normalized and labeled accelerometer data corresponding to one trial of a given participant (P01\_290416). Each data cluster corresponding to an activity label has a different color on the plot.}
    \label{fig:acc_norm}
\end{figure}

\begin{figure}
    \centering
    \includegraphics[width=5in]{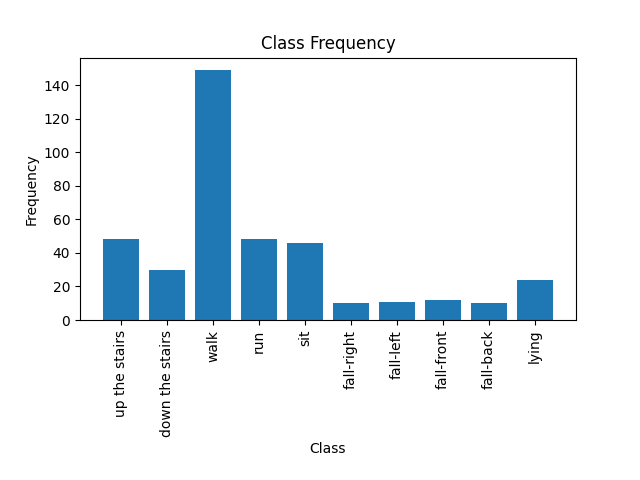}
    \caption{Bar chart of class distribution of all trials of a given participant (P01\_290416), providing a strong visual indication of class imbalance in the dataset D.}
    \label{fig:class_dist}
\end{figure}

Let $A_x$, $A_y$ and $A_z$ denote respectively the accelerometer data  along the $x$-axis, $y$-axis and $z$-axis.  We also compute the Euclidean norm $A_n$ of the three-axis acceleration signal: (Figure~\ref{fig:acc_norm}).
\begin{equation}
    A_n = \sqrt{A_x^2 + A_y^2 + A_z^2 }
\end{equation}
The data is finally represented in a csv file (the dataset $D$) in a structured format, with the following attributes,
\begin{itemize}
    \item index: line number in the csv file ($N= 3,5$ millions data points)
    \item participant number, in $\{1, \ldots, 44\}$
    \item participant reference: Nomenclature as saved in the LIO server $PXX\_ddmmYY$
    \item trial number, in $\{1, \ldots, 6\}$
    \item timestamp $YYYY-mm-dd \ HH:MM:SS.FFF$ (format in milliseconds)
    \item $A_x$: data from the $x$-axis accelerometer sensor
    \item $A_y$: data from the $y$-axis accelerometer sensor
    \item $A_z$: data from the $z$-axis accelerometer sensor
    \item $A_n$: euclidean norm
    \item activity name
\end{itemize}
Since this dataset represents activities in real contexts, class imbalances occur. For example, as it can be seen in Figure~\ref{fig:class_dist}, there are more instances of walking than other activities.
\paragraph{Validation}
\label{parag:val}
The validation is performed using leave-one-subject-out cross validation of the dataset~\cite{Arlot2009}, using the scikit-learn package. Dehghani et al.~\cite{dehghani2019} recommended its use for performance evaluation in human activity recognition. The dataset is split according to the number of subjects in the dataset. 
That is, in each fold the model is trained on all the subjects except one, which is used for testing. In this case, the number of folds is equal to the number of subjects in the dataset~\cite{dehghani2019}. 
In order to have an accurate estimation of the proposed method performance, this procedure is repeated until all the subjects have been used as test datasets. Obviously, activity patterns are subject-dependent. That is, the performed activities vary considerably among subjects according to their personal styles and anthropometry, which we refer to as inter-subject variability. Moreover, the way a given activity is performed by a given subject at different times may show variations, which we refer to as intra-subject variability. In our work, the intra-subject variability is considered when the participants are instructed to repeat the sequence of activities a given number of times (i.e., the number of trials). Using the leave-one-subject-out strategy guarantees that data from the same subject is either present in the train set or in the test set. This is a subject independent (inter-subject) validation technique to estimate generalization capabilities of learning techniques, and also allows performance analysis results per subject. 
\paragraph{Segmentation}
Here, the input data ($A_x$, $A_y$, $A_z$, and $A_n$) are chronologically sequenced to form multivariate time series signals (of length $N$). We segment these raw and continuous data flow signals, using a $1$-s fixed-size overlapping sliding window (FOSW), into $K$ fixed-length segments, with a $50\%$ overlap, in order to search of useful patterns in the time series. The window size T=1s corresponds to $64$ data points, and is chosen according to the accelerometer frequency (64 Hz).
The segments are stored in a three-dimensional tensor of size $K \times T \times m$, where $K$ is the number of segments, $T$ is the time step (window size), and $m$ is the number of multivariate time series. The three dimensional tensor is used as the input of the classifier.
The tensor-based approach allows to reduce the computational
cost when dealing with big data. Here $m = 4$, which corresponds to $A_x$, $A_y$, $A_z$ and $A_n$.\\
In temporal segmentation, the problem of impure segments may occur. Since for the transition points between activities, a segment is composed of more than one label~\cite{Liono2016}.
To overcome the multi-class segment problem, ambiguous segments, that contain more than a label, are discarded. 
We also note that we have performed the train/test split before the segmentation, in order to avoid train and test data overlapping.

%% file: Methods.tex
\label{methods}
\subsection{Overview of the proposed architecture}
\begin{figure}
\centering
\includegraphics[width=5in]{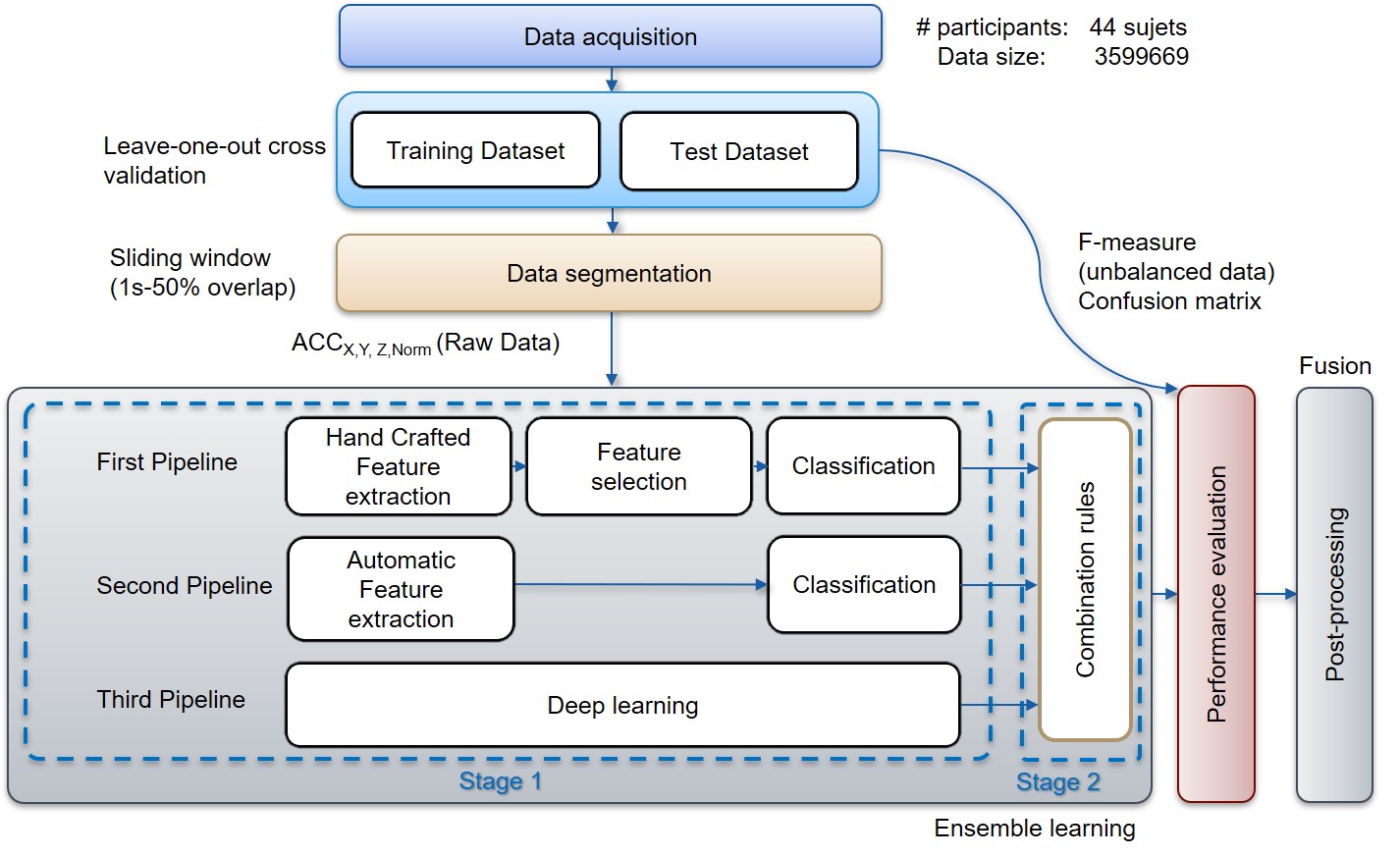}
\caption{Overview of the proposed algorithm architecture.\label{figflow}}
\end{figure}
We are considering a multi-class pattern classification task, where the input (i.e., multivariate time series signals) is to be classified into one, and only one, of the $l$ non-overlapping classes referring to the considered human physical activities in Section~\ref{research_protocol}. \\
The main contribution of this study is the development of a reference architecture design based on an ensemble learning system that uses both machine learning and deep learning techniques on a large dataset in order to achieve the highest generalization performance. Moreover, a parallel training approach for the proposed architecture is developed to accelerate the cross validation procedure by running multiple independent training tasks on multiple cores or processors in parallel. In other words, we build an ensemble learning system with an efficient cross validation implementation using parallel processing in order to balance the generalization performance and to reduce the computational cost of the system. Figure~\ref{figflow} shows the proposed ensemble learning architecture. 
The proposed ensemble learning architecture combines three data representation-based approaches. In the remainder of the paper, the process of tying together different learning algorithms is referred to as a pipeline. The first pipeline is based on a handcrafted-feature engineering approach. This approach involves handcrafted feature extraction techniques, feature selection and machine learning algorithms. The second pipeline is based on an automatic feature extraction approach. This approach involves automatic feature extraction and machine learning algorithms. The third pipeline is based on a feature learning approach. This approach involves deep learning algorithms. Afterward, a \textbf{classifier fusion} method is used to yield the final prediction results~\cite{Ruta2000}, in order to combine the resulting three pipeline decisions.\\

Moreover, the proposed architecture deals with multivariate accelerometer time series data. Hence, in each stand-alone classifier, we separate multivariate time series into univariate ones ($A_x$, $A_y$, $A_z$ and $A_n$ separately), and we perform either feature engineering or feature learning on each univariate time series individually. \textbf{Axis-based feature fusion} concatenates different resulting features from all the univariate time series input before the classification~\cite{Fu2008}.\\

\subsubsection{First pipeline}
In this approach, we apply a support vector machine algorithm for the classification of human physical activities using
time and frequency-domain features and the reliefF-based feature selection algorithm.
\paragraph{Handcrafted feature extraction}
For each feature univariate time series, commonly used time and frequency domain metrics are computed for each segment, to extract basic signal information from raw accelerometer data~\cite{Figo2010}.
\begin{itemize}
    \item Time domain metrics: mean, variance, standard deviation, maximum, minimum, Root Mean Square (RMS), kurtosis, skewness, euclidean norm and l$1$-norm.
    \item Frequency domain metrics: energy and  maximum magnitude of the Fast Fourier Transform (FFT).
\end{itemize}
Moreover, the cross-correlation is computed in a pairwise fashion between each two features (i.e., ($A_x,A_y$), ($A_x,A_z$), ($A_y,A_z$)) $\dots$, on each segment.\\
The handcrafted feature extraction procedure leads to a total of $55$ time-frequency features per segment that are z-score normalized.  
\paragraph{Feature selection using ReliefF}
The resulting features are ranked using the ReliefF feature selection algorithm~\cite{Kononenko1997}. The algorithm is based on computing the importance of the features by randomly choosing a given number of instances in the dataset (this number is a user-defined parameter) and searching for its two nearest neighbors: one from the same class and the other from a different class. 
\paragraph{Multi-class Support Vector Machine (SVM) classifier}
The next step in this classification pipeline, is the use of the Support Vector Machine (SVM), which is a supervised machine learning algorithm~\cite{Hearst1998}. For the binary classification SVM, the idea is to construct a hyperplane to separate the two classes so that the distance between the hyperplane and the sample points is maximal. \\
One approach to solve the $l$-class SVM ($l > 2$) problem is to consider the problem as $l$ binary SVMs. This approach is called one-vs-all SVM. Another approach consists in considering the problem as $l \times (l -1)/2$ binary classifiers using all the binary pair-wise combinations of the $l$ classes. This approach is called one-vs-one SVM. In this study, we considered the one-vs-all SVM.
The kernel (i.e., the type of hyperplane used to separate the data), gamma (i.e., a parameter for non linear hyperplanes), and C (i.e., the penalty parameter of the error term) are the hyperparameters of SVM, which need to be fine-tuned.

\subsubsection{Second pipeline}
For this pipeline, we use the $K$-nearest neighbor algorithm, trained for the classification of physical human activities using the linear discriminant analysis as an automatic feature extraction method.
\paragraph{Automatic feature extraction using Linear Discriminant Analysis (LDA)}
Automatic feature extraction consists of mapping the raw data onto a lower-dimensional space. LDA is a supervised feature extraction algorithm, that takes the labels of the training data into account, and aims at maximizing the class discrimination on the projected space~\cite{xanthopoulos2013linear}. The LDA algorithm is applied to each feature (i.e., the univariate time series). Then, the feature fusion concatenates the different transformed features to use them as an input for the K-nearest neighbor (KNN) classifier.

\paragraph{K-Nearest Neighbor (KNN) classifier}
The principle of the KNN algorithm is to find the $k$ closest in distance training samples to the current sample, and predict its label based on these $k$ neighbors~\cite{peterson2009k}. Generally, the value of $k$ is specified by the user. The distance can, in general, be any metric measure. We note however that the standard Euclidean distance is the most common choice. The optimal choice of $k$ is highly data-dependent, and requires to be fine-tuned.
\subsubsection{Third pipeline}
We adopt a convolutional neural network architecture for multivariate time series trained for both feature extraction and classification.
\paragraph{Convolutional Neural Network (CNN) for multivariate time series}
\begin{figure}
    \centering
    \includegraphics[width=5in]{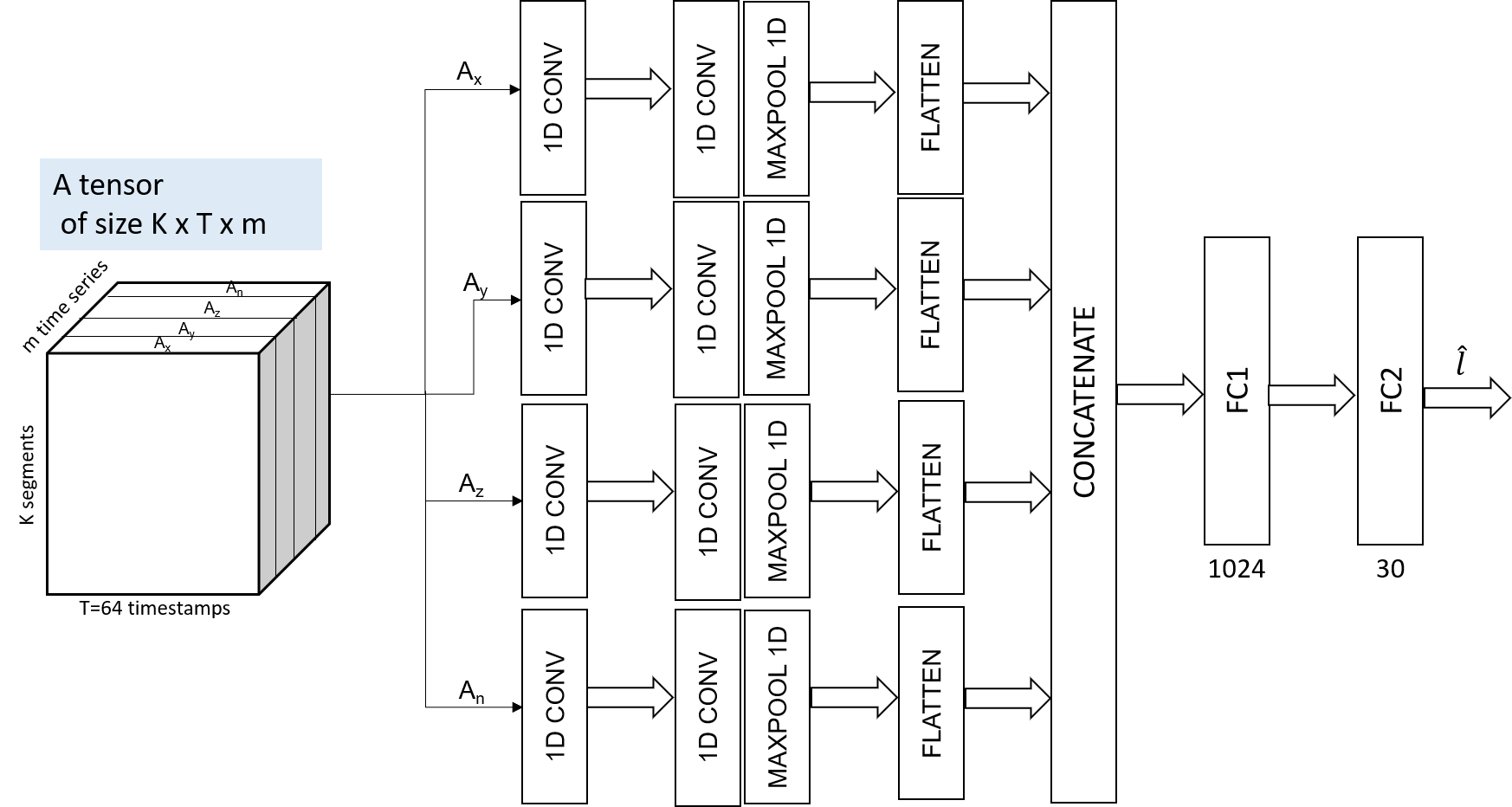}
    \caption{The three dimensional tensor representation of the multivarite time series data is the input of the convolutional neural network pipeline of our architecture. }
    \label{fig:cnn}
\end{figure}
Let's first recall that a deep neural network has an input layer, an output layer and more than two hidden layers. A layer is a collection of neurons. A neuron takes a group of weighted inputs, applies a non linear activation function, and returns an output.
\begin{itemize}
    \item The input layer has $T \times m$  neurons.
    \item Hidden layers of a deep network are designed to learn the hierarchical feature representations of the data. During the training, a set of hyper-parameters is optimized, and the weights are initialized randomly~\cite{LeCun2012}. By gradient descent, the weights are updated using the back propagation algorithm, in a way that minimizes the cost function on the training set. The choice of the model, the architecture and the cost function is crucial to obtain a network that generalizes well, and is in general problem and data dependent. 
    \item The output layer has $l$ neurons, which corresponds to the multi-class classification problem in this application. 
\end{itemize}

In this work, we train a CNN-based architecture. We recall that CNNs can capture the local connections of multimodal sensory data~\cite{hammerla2016}. CNN combines three architectural ideas: local receptive fields, shared weights, and pooling, and it is based on two building blocks:
\begin{itemize}
    \item The convolution block which is composed of the convolution layer and the pooling layer. These two layers form the essential components of the feature extractor, which learns the features from the raw data automatically (feature learning).\\
The convolutional layer implements the receptive field and shared weight concepts. Neurons in the convolutional layers are locally connected to neurons inside its receptive field in the previous layer. Neurons in a given layer are organized in planes where all the neurons share the same set of weights (also called filters or kernels). The set of outputs of the neurons in such a plane is called a feature map. The number of feature maps are the same as the number of filters. A pooling layer performs either an average sub-sampling (mean-pooling) or a maximum sub-sampling (max-pooling).
For a time series, the pooling layers simply reduce the length, and thus the resolution, of the feature maps.
\item The fully connected block which performs the classification based on the learned features from the convolutional blocks.
\end{itemize}
The different hyper-parameters of CNN are the optimization algorithm (momentum), the number of epochs, the number of layers, the number of filters, the filter size, the activation function, the cost function, the batch size and the weight initialization~\cite{LeCun1998}.\/
We apply a CNN model to the multivariate time series classification task at hand. First, we separate multivariate time series into univariate ones and perform feature learning on each univariate time series individually. Then, we concatenate the resulting features at the end of the feature learning step to do the classification.
Our CNN-based model has three layers including two convolutional blocks, and two fully-connected layers.%, as depicted in figure~\ref{fig:cnn}. 
That is, for each feature, the input (i.e., univariate time series) is fed into a one-stage feature extractor (the convolutional block). The convolutional block consists of a convolution layer with three filters, a filter size of $20$, a REctified Linear Units (ReLU) activation layer and a max pooling layer with pooling size of $3$. At the end of the feature extraction step, we flatten the feature maps of each univariate time series and combine them as the input of subsequent fully connected block for classification. The fully connected block consists of two fully connected layers with $1024$ and $30$ neurons, respectively, with a ReLU activation function \cite{zeng2014} (Figure \ref{fig:cnn}.
\subsubsection{Fusion stage}
Each pipeline computes a prediction vector for the same training dataset. The final decision is made by a combination of the three pipeline decisions (prediction vectors) using a decision rule (majority voting)~\cite{Lam1997} to produce the final classification result.

Here, we predict the final class label $\hat{l}$ based on the majority (plurality) voting of each classifier $l_i$:\\
$$\hat{l} = mode\{l_1,l_2,l_3\}$$
i.e., the final class label $\hat{l}$ corresponds to the class label that has been predicted most frequently by the three used classifiers within the three pipelines of our architecture. 

\subsubsection{Computational optimization}
\label{computational-opti}
One of the issue to consider when designing and developing machine and deep learning applications is the computational cost. In fact, on one hand the more data we consider the more robust the application is but on the other hand the more time consuming the application is as well, regarding both the data processing and the best model building and selection. To overcome such an issue, one should consider a more efficient usage of the computing resources by considering parallel and/or distributed programs. For example, the leave-one-subject-out cross validation that we use in our architecture is computationally expensive. However, since the tasks performed on each fold are completely independent from the rest of the remaining folds, we are running as many folds as the core number of the target architecture in parallel. To do so, we use the multiprocessing parallel python package~\cite{multiprocessing}. We also note that in order to further enhance the performance of our system, we use a GPU accelerator to run the mutivariate CNN model from the python keras package with tensorflow backend~\cite{tensorflow2015-whitepaper}. To ensure a good balance between the generalization performance of the model and its computational performance, the different hyperparameters of the deep learning model such as the batch\_size should be carefully chosen~\cite{smith2018}.
In this application, we first use the python multiprocessing package which allows, just by adding a few lines of code, to run up to $5 \times$ faster comparing to the sequential code using a machine with $8$ cores and $1$ NVIDIA TITAN RTX GPU. A more detailed description of the machine will be given in the following section.

%% file: Results.tex
\section{Results}
\label{results}
\subsection{Experimental Design}
\paragraph{Weighting Imbalanced Classes}
 To handle the class imbalance problem, we do not include any oversampling but an algorithm-level method. That is, class weight is added that automatically assigns higher weights to the minority classes in the learning process, in order to reduce bias towards the majority group~\cite{ Krawczyk2016, Johnson2019}. 
\paragraph{Multi-class Performance Measures}
The result of an $l$-class classification can be visualized in a confusion matrix of size $l\times l$. Each row in a confusion matrix represents an actual class, while each column represents a predicted class. By definition, each entry $C_{ij}$ in a confusion matrix $\mathbf{C}$ denotes the number of observations (segments) from class $i$ predicted to be of class $j$.\\
The recall, precision and F1-score are used as performance metrics to evaluate the correctness of a classification. The F1-score metric is particularly interesting for its robustness to class imbalance~\cite{Sokolova2009}.
The recall ($R_i$), precision ($P_i$) and F1-measure ($F1_i$) for class $i$ in a multiclass problem can be defined by the following equations,\\
$$P_i = \frac{TP_i}{TP_i + FP_i},~~R_i = \frac{TP_i}{TP_i + FN_i},~~ \textrm{and}~~~F1_i = 2 * \frac{P_i * R_i}{P_i + R_i}$$
~~ \textrm{where}~~~ $TP_i$ is the number of objects from class $i$ assigned correctly to class $i$, $FP_i$ is the number of objects that do not belong to class $i$ but are assigned to class $i$, and $FN_i$ is the number of objects from class $i$ predicted to another class.

The quality of the overall classification is usually assessed in two ways: macro-averaging and micro-averaging. The first computes the measure separately for each class and then takes their unweighted mean. A weighted average could be computed by support (i.e., the number of true instances for each label) to account for class imbalance. The second calculates the measure globally by counting the total true positives, false negatives and false positives.\\
In the following equations, $\kappa$ and $\mu$ indices refer to the macro- and micro-averaging, respectively. $P$, $R$, and $F1$ are the total precision, recall and F1 measures~\cite{Sokolova2009}.
\begin{center}

$$P_\kappa = \frac{1}{K} \sum_{i=1}^{K} P_i, ~~R_\kappa = \frac{1}{K} \sum_{i=1}^{K} R_i, ~~ \textrm{and}~~~F1_\kappa = 2 * \frac{P_\kappa * R_\kappa}{P_\kappa + R_\kappa}$$
\end{center}
\begin{center}
$$P_\mu = \frac{\sum_{i=1}^{K} TP_i}{\sum_{i=1}^{K} (TP_i + FP_i)},  ~~R_\mu = \frac{\sum_{i=1}^{K} TP_i}{\sum_{i=1}^{K} (TP_i + FN_i)}~~ \textrm{and}~~~F1_\mu = 2 * \frac{P_\mu * R_\mu}{P_\mu + R_\mu}$$
\end{center}
\subsection{Experimental results}
The classification is performed using a leave-one-subject-out cross-validation as detailed in section~\ref{parag:val}. Instead of averaging the performance measures of each holdout fold, predictions are computed and stored in a list. Then, at the end of the run, the predictions are compared to the expected values for each holdout test set and a single performance measure is reported. 

\paragraph{Classification result using the handcrafted feature engineering based approach}
\begin{table}
\caption{Classification report using the handcrafted feature engineering based approach}
\label{table:RF_SVM}
\centering
\begin{tabular}{c | c c c r}
Class & Precision & Recall & F1-score & Support\\
\midrule
up the stairs & 0.87 & 0.96 & 0.91 & 268\\
down the stairs & 0.99 & 0.88 & 0.93 & 199\\
walk & 0.98 & 0.94 & 0.96 & 702\\
run & 0.94 & 1.00 & 0.97 & 372\\
sit & 0.77 & 1.00 & 0.87 & 345\\
fall-right & 1.00 & 1.00 & 1.00 & 5\\
fall-left & 1.00 & 1.00 & 1.00 & 6\\
fall-front & 1.00 & 0.67 & 0.80 & 6\\
fall-back & 0.83 & 1.00 & 0.91 & 10\\
lying & 1.00 & 0.73 & 0.85 & 394\\
\midrule
macro avg & 0.94 & 0.92 & 0.92 & 2307\\
weighted avg & 0.93 & 0.92 & 0.92 & 2307\\
\end{tabular}
\end{table}
\begin{figure}[!htbp]
    \centering
    \includegraphics[scale=0.8]{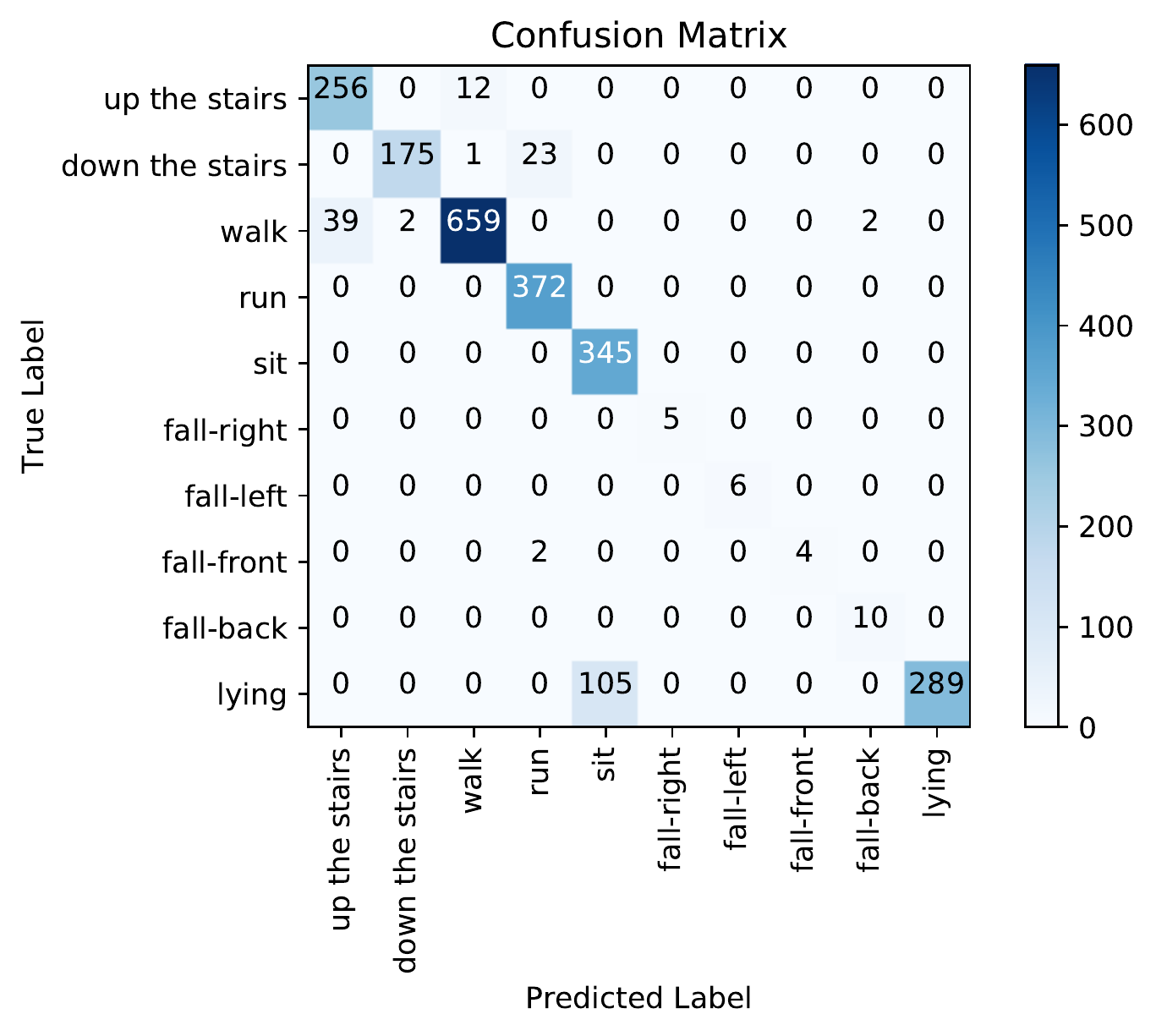}
    \caption{Confusion matrix using the handcrafted feature  engineering based approach.}
    \label{fig:RF_SVM}
\end{figure}

Figure~\ref{fig:RF_SVM} presents the confusion matrix for inter-subject activity recognition obtained using the adopted method with time and frequency domain features, ReliefF feature selection, and SVM classifier (first pipeline). Table~\ref{table:RF_SVM} shows medium to high precision scores (from $77\%$ to $100\%$), medium to high recall scores (from $67\%$ to $100\%$), and high F1-scores (from $80\%$ to $100\%$) on each activity class. The overall weighted averaged precision, recall, and F1-score across all activities are $93\%$, $92\%$, and $92\%$, respectively.\\
If we examine the recognition performance for each activity individually, walking up the stairs is confused with walking ($0.04\%$), walking down the stairs is confused with running ($0.11\%$), walking is confused with walking up the stairs ($0.26\%$), and lying is confused with sitting ($0.27\%$).
\paragraph{Classification result using the automatic feature extraction based approach}
\begin{table}
\caption{Classification report using the automatic feature extraction based approach.}
\label{table:LDA_KNN}
\centering
\begin{tabular}{c | c c c r}
Class & Precision & Recall & F-score & Support\\
\midrule
up the stairs & 0.91 & 0.81 & 0.85 & 268\\
down the stairs & 0.77 & 0.35 & 0.48 & 199\\
walk & 0.77 & 0.96 & 0.86 & 702\\
run & 0.97 & 0.82 & 0.89 & 372\\
sit & 0.94 & 1.00 & 0.97 & 345\\
fall-right & 1.00 & 0.80 & 0.89 & 5\\
fall-left & 1.00 & 1.00 & 1.00 & 6\\
fall-front & 1.00 & 0.83 & 0.91 & 6\\
fall-back & 1.00 & 0.70 & 0.82 & 10\\
lying & 0.99 & 1.00 & 1.00 & 394\\
\midrule
macro avg & 0.93 & 0.83 & 0.87 & 2307\\
weighted avg & 0.88 & 0.88 & 0.87 & 2307\\
\end{tabular}
\end{table}

\begin{figure}[!htbpt]
    \centering
    \includegraphics[scale=0.8]{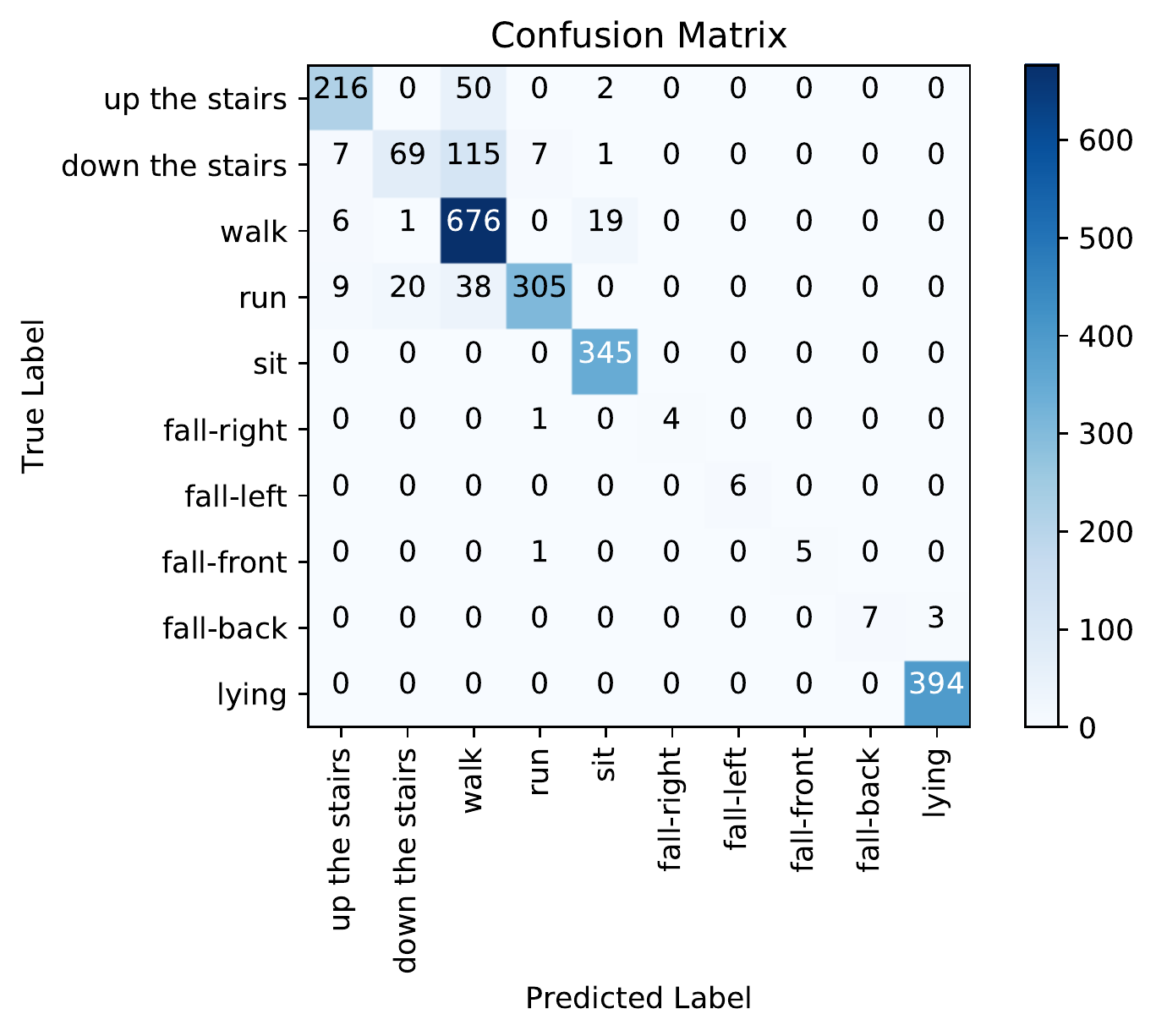}
    \caption{Confusion matrix using the automatic feature extraction based approach}
    \label{fig:LDA_KNN}
\end{figure}
Figure~\ref{fig:LDA_KNN} presents the confusion matrix for inter-subject activity recognition obtained using the adopted automatic feature extraction based on LDA, and KNN classifier (second pipeline). Table~\ref{table:LDA_KNN} shows medium to high precision scores (from $77\%$ to $100\%$), low to high recall scores (from $35\%$ to $100\%$), and low to high F1-scores (from $48\%$ to $100\%$) on each activity class.
 
The overall weighted averaged precision, recall and F1-score, across all activities, are $88\%$, $88\%$, and $87\%$, respectively.
If we examine the recognition performance for each activity individually, we can see that again, walking up the stairs is confused with walking ($0.18\%$), walking down the stairs is confused with walking ($0.77\%$), walking is confused with sitting ($0.02\%$), running is confused with walking up the stairs ($0.02\%$), down the stairs ($0.05\%$) and walking ($0.10\%$). We notice that this approach performs worse than the previous approach on dynamic activities (walking, walking up and down the stairs and running). However, it performs better on static activities (sitting and lying).

\paragraph{Classification result using feature learning based approach}
\begin{table}[!htbp]
\caption{Classification report using feature learning based approach}
\label{table:MCNN}
\centering
\begin{tabular}{c | c c c r}
Class & Precision & Recall & F-score & Support\\
\midrule
up the stairs & 0.99 & 0.94 & 0.97 & 268\\
down the stairs & 0.99 & 0.96 & 0.98 & 199\\
walk & 0.98 & 1.00 & 0.99 & 702\\
run & 0.98 & 1.00 & 0.99 & 372\\
sit & 1.00 & 1.00 & 1.00 & 345\\
fall-right & 1.00 & 1.00 & 1.00 & 5\\
fall-left & 0.86 & 1.00 & 0.92 & 6\\
fall-front & 0.86 & 1.00 & 0.92 & 6\\
fall-back & 1.00 & 0.90 & 0.95 & 10\\
lying & 1.00 & 1.00 & 1.00 & 394\\
\midrule
macro avg & 0.97 & 0.98 & 0.97 & 2307\\
weighted avg & 0.99 & 0.99 & 0.99 & 2307\\
\end{tabular}
\end{table}
\begin{figure}[!htbp]
    \centering
    \includegraphics[scale=0.8]{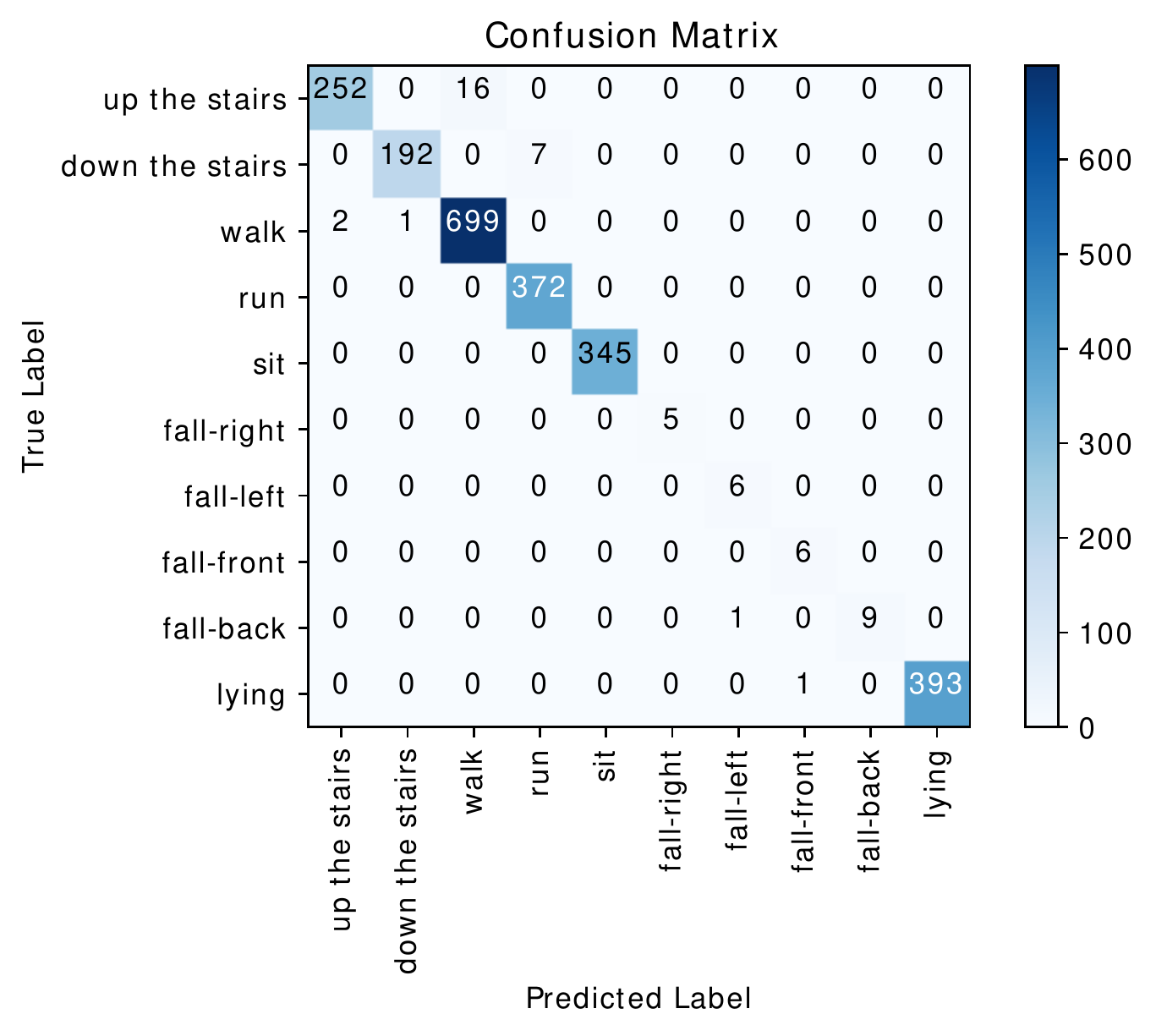}
    \caption{Confusion matrix using feature learning based approach.}
    \label{fig:MCNN}
\end{figure}
Figure~\ref{fig:MCNN} presents the confusion matrix for inter-subject activity recognition obtained using the adopted multivariate CNN classifier (third pipeline). Table~\ref{table:MCNN} shows high precision (from $86\%$ to $100\%$), high recall (from $90\%$ to $100\%$), and high F1-score (from $92\%$ to $100\%$) on each activity class. The overall weighted averaged precision, recall, and F1-score across all activities are all equal to $99\%$.
If we examine the recognition performance for each activity individually, walking up the stairs is confused with walking ($0.05\%$), walking down the stairs is confused with running ($0.03\%$), walking is confused with walking up the stairs ($0.002\%$) and down the stairs ($0.001\%$). We notice that this approach performs better than the two previous approaches on recognizing both dynamic and static activities. 

\begin{figure}[!htbp]
    \centering
    \includegraphics[scale=0.8]{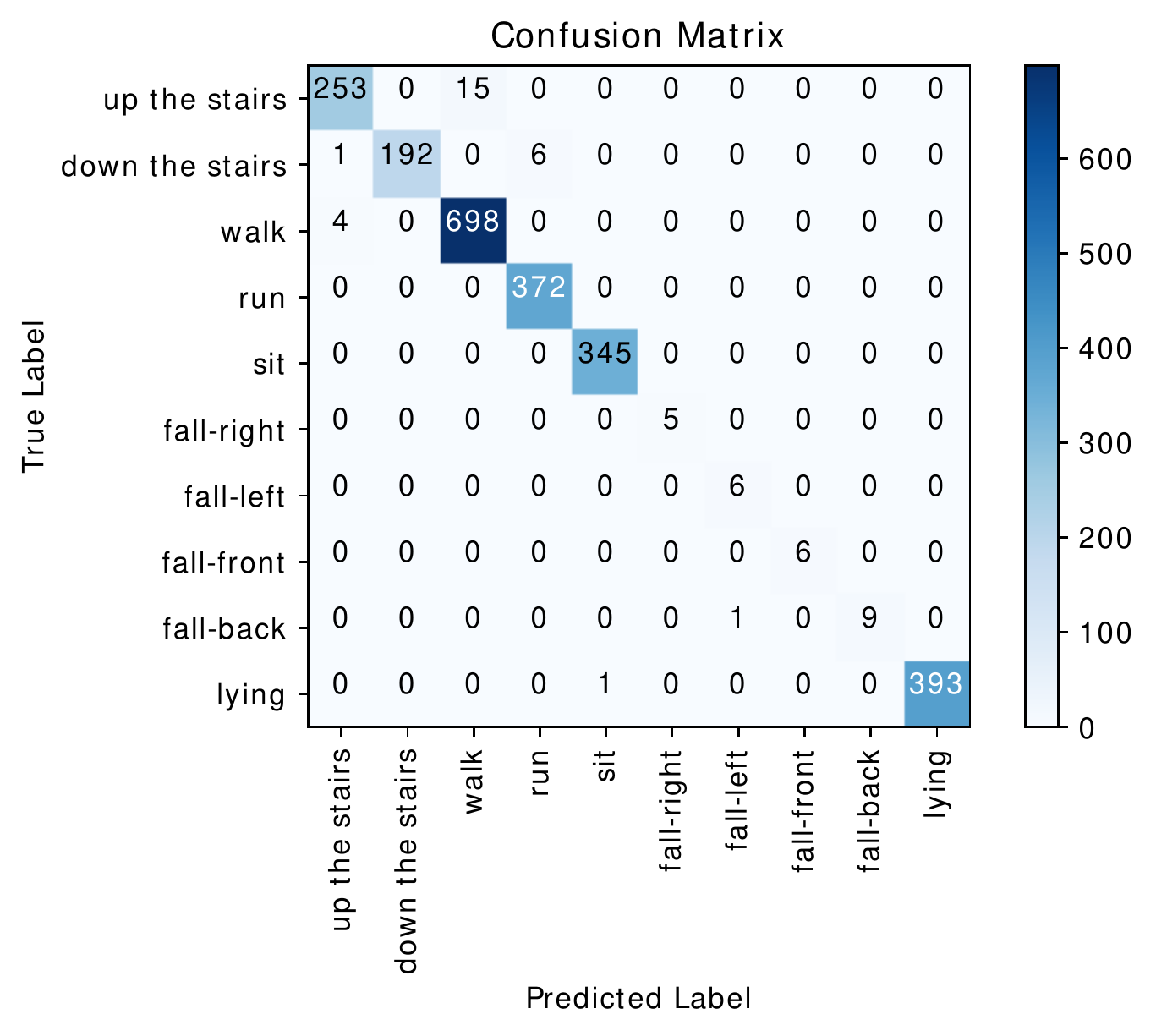}
    \caption{Confusion matrix using the ensemble learning based approach}
    \label{fig:ens_lern}
\end{figure}

\paragraph{Classification results using the ensemble learning based approach}
Figure~\ref{fig:ens_lern} presents the confusion matrix for inter-subject activity recognition obtained using the proposed ensemble learning approach, that is combining the results of the three previous pipelines. Table~\ref{table:ens_lern}
shows high precision (from $86\%$ to $100\%$), high recall (from $90\%$ to $100\%$) and high F1-score (from $92\%$ to $100\%$) on each activity class. The overall weighted averaged precision, recall and F1-score across all activities are about $99\%$. If we examine the recognition performance for each activity individually, walking up the stairs is confused with walking ($0.05\%$), walking down the stairs is confused with running ($0.03\%$), walking is confused with walking up the stairs ($0.002\%$). The proposed ensemble learning approach performed extremely well in recognizing the ten different activities compared to the three previous approaches considered individually.
\begin{table}
\caption{Classification report using the ensemble learning based approach}
\label{table:ens_lern}
\centering
\begin{tabular}{c | c c c r}
Class & Precision & Recall & F1-score & Support\\
\midrule
up the stairs & 0.98 & 0.94 & 0.96 & 268\\
down the stairs & 1.00 & 0.96 & 0.98 & 199\\
walk & 0.98 & 0.99 & 0.99 & 702\\
run & 0.98 & 1.00 & 0.99 & 372\\
sit & 1.00 & 1.00 & 1.00 & 345\\
fall-right & 1.00 & 1.00 & 1.00 & 5\\
fall-left & 0.86 & 1.00 & 0.92 & 6\\
fall-front & 1.00 & 1.00 & 1.00 & 6\\
fall-back & 1.00 & 0.90 & 0.95 & 10\\
lying & 1.00 & 1.00 & 1.00 & 394\\
\midrule
macro avg & 0.98 & 0.98 & 0.98 & 2307\\
weighted avg & 0.99 & 0.99 & 0.99 & 2307\\
\end{tabular}
\end{table}

\subsection{Discussion of the recognition rate results}
In this paper, we focus on the recognition of three types of human physical activities (static activity, dynamic activity, and hazardous situations such as falling). Each of the three tested approaches has its own strengths and weaknesses on classifying different type of activities. Sitting and lying are static activities. Dynamic activities include walking, walking up and down the stairs and running. Concerning dynamic activities, we recall that walking generates a periodic pattern, running implies a motion similar to walking, but executed faster. Looking at Figure~\ref{xyzplots}, a periodic pattern can also be noticed for running, with a shorter time difference between periods compared to walking. Some false positives and negatives showed up when trying to distinguish and recognize these two activities using the automatic feature extraction based approach.
Moreover, walking and walking up and down the stairs are almost similar (see Figure~\ref{xyzplots}), and can be performed in many ways. Thus, they are easily confused. However, the proposed ensemble learning model performed extremely well when identifying the different dynamic activities compared to standalone approaches.\\
Concerning static activities, We note that lying down and sitting are confused in the handcrafted feature engineering approach since the orientation of waist-worn accelerometer is similar for both activities.\\
As described in the confusion matrices~\ref{fig:RF_SVM},~\ref{fig:LDA_KNN},\ref{fig:MCNN} and~\ref{fig:ens_lern}, we can see that for the four fall classes, labeled observations are scarce compared to the other classes such as walking and running for example (see Figure~\ref{fig:class_dist}). This is explained by the short duration of the falls. The consideration of this class in our system aims at detecting a fall if it occurs during the rehabilitation process. We recall however, as previously explained in the protocol that the falls we are considering here are simulated and may be different from an actual fall. We also consider four classes of falling to show that our system is able to capture the direction of the fall, which is very important since falling front or back have in general a more important impact.

\subsection{Performance speed analysis}
Here, we present the computational cost of the parallel HAR system we developed. We run our experiments on a machine with an intel Core $i7-9700$ processor having $8$ cores and a $64GB$ RAM connected to a GPU accelerator of type NVIDIA TITAN RTX. One of the bottleneck in terms of running time in our system corresponds to the segmentation stage. In fact, running a sequential segmentation (using only one core of the cpu) relative to the $44$ participants based on the leave-one-subject-out validation technique takes around $6.5$ days. However, when we run the segmentation in parallel making full usage of the $8$ cores of our machine via a parallel implementation based on the multiprocessing python package, it takes around $32$ hours, which is almost $5\times$ faster than the sequential version. As described in Section~\ref{computational-opti}, each fold training relative to one of the train/test split is run as an independent task. Thus, a task queue is formed and the different tasks are assigned to the $8$ cores of the machine. Each task is a succession of the three pipelines of our learning models applied to the considered fold. Here again, we use the multiprocessing python package, and we obtain a speed up of almost $3$ compared to the sequential implementation. We also note that within each task, the third pipeline, which corresponds to the multivariate CNN method is run on the GPU accelerator, which takes around $15$ seconds, that is $80\times$ faster than a purely cpu version, which takes around $20$ minutes. We mention here that giving that each task is running on an independent data subset, our performance results regarding the computational time could be further improved by using a larger number of cores and/or processors.

%% file: Conclusion.tex
\section{Conclusions}
\label{conclusion}
The main goal of this research work is the development of an adherence measurement system that is completely objective, precise and efficient in term of computational resource usage. This requires the development of a recognition system for human physical activity, which is the part we focus on in this paper. In the following, we summarize our contributions and the limitations of the proposed method. Then, we present some possible perspectives and insights.\\
The first part of this research is the acquisition of a large multi-modal dataset based on wearable sensor vest that captures cardiac, respiratory, and accelerometer data from healthy and young volunteers. Each participant undergoes a sequence of ten real-life physical activities including static and dynamic activities, that are likely to occur during a patient rehabilitation protocol~\cite{Brewer1998}. The purpose of this data acquisition is to establish a proof-of-concept that the recorded acceleration data from the waist-worn accelerometer during physical activity, could be modeled for a HAR system using learning techniques for the purpose of health monitoring in an upcoming stage of a more global research project. \\
The second part is the development of a recognition system for these human physical activities using accelerometer data collected from a waist-mounted accelerometer in the vest. We note that the model may be extended by some other activities. Later, the pre-trained HAR system should be able to classify activities of patients undergoing a cardiac rehabilitation program. 
The developed HAR system is based on an ensemble learning architecture that combines different data representation-based classifiers (feature engineering, automatic feature extraction and feature learning). The output of these classifiers are combined to improve the classification performance (minimizing false positives and false negatives). An inter-subject validation strategy (leave-one-subject-out cross validation) is used to have a realistic estimate of the performance generalization of each classifier independently, as well as the ensemble architecture. However, classifier ensembles combined to the leave-one-subject-out validation technique are clearly more expensive on large datasets, computationally speaking, as they require several models to be trained.  
Hence, we propose a parallel implementation for our architecture to accelerate the cross validation procedure by running multiple fold training simultaneously. We also enhance the computation relative to each fold by the use of a GPU accelerator. Another advantage of our implementation is the fact that the segmented multivariate time series are stored into a fourth-order $3D$ tensor. Finally, we demonstrate that forms of locomotion such as walking, running, and climbing the stairs, and postures such as sitting and lying down, as well as some hazardous situations such as falling can be recognized at up to $99\%$ recognition rate using waist-worn accelerator.\\
In this work, we focus on classifying activities collected in a laboratory environment. However, for the more general research project, it is important to train and test activity recognition systems on data collected under naturalistic circumstances, with patients undergoing a rehabilitation program. We note that we planned to collect data from a pathological population following a cardiac rehabilitation program in collaboration with the Centre de cardiologie préventive (CCP) in the Centre Hospitalier de l'Universit\'{e} de Montr\'{e}al (CHUM), to rigorously validate the efficiency of the proposed HAR system. \\
Nevertheless, this study has some limitations. For example, ground truth data has been semi-automatically annotated. The annotation procedure may be fine-tuned, and a real-time automatic annotation may be investigated in order to increase the volume of data collected.
Moreover, we do not take into account transitional activities neither in the segmentation nor in the classification processes. Transition-aware and activity-based segmentation approaches could be investigated.\\
In the following, we enumerate five possible directions for future work.
The first is to tune the hyperparameters of the used learning algorithms in this paper in order to get the best performance on our dataset. The second is to investigate open-source feature engineering libraries (for instance, tsfel~\cite{BARANDAS2020} and tsfresh~\cite{CHRIST2018}) for time series, to capture as much as possible of discriminative signal characteristics of human physical activities. The third is to understand the learned features by each proposed approach (feature engineering, automatic feature extraction and feature learning) by applying a Class Activation Map (CAM)~\cite{Zhou2016} for example, then comparing the corresponding output feature vectors. The fourth is to smooth and fuse the predicted activity labels of consecutive segments. Thus, identifying the start and end points of each activity is very useful for physical rehabilitation, to have better 
information about activity transitions and the duration of each activity. Finally, each pipeline in the proposed ensemble learning architecture could be extended by evaluating more than one standard learning algorithm belonging to each approach, on the data at hand, and fusing their predictions to achieve a better recognition rate.
We should note that the adopted learning algorithms used at each pipeline serve as a proof of concept of the proposed ensemble learning architecture. \\

%% file: Acknowledgment.tex
\paragraph{Author contributions:}
Conceptualization, M.A. and A.K.; methodology, M.A. and A.K.; software, M.A., A.K. and Y.O.; validation, M.A. and A.K.; formal analysis,  M.A. and A.K.; investigation,  M.A. and A.K.; resources, X.X.; data curation, Y.O., H.W. and S.C.; writing—original draft preparation, M.A. and A.K.; writing—review and editing, M.A. and A.K.; visualization, M.A. and A.K.; supervision, N.M., A.M. and A.B.B.; project administration, M.A.; funding acquisition, N.M. All authors have read and agreed to the published version of the manuscript.
\paragraph{Acknowledgements:} We would thank the participants for taking part in this study.
\paragraph{Funding:} This research was funded by the Canada Research Chair on Biomedical Data Mining (950-231214), the discovery grants program of the natural sciences and engineering research council of canada (NSERC), and the collaborative research and development grant (CRD) of the NSERC, where Carré Technologies Inc. is the industrial partner, and Prompt is the financial partner.